\def\year{2019}\relax
\documentclass[letterpaper]{article} 
\usepackage{aaai19}  
\usepackage{times}  
\usepackage{helvet}  
\usepackage{courier}  
\usepackage{url}  
\usepackage{graphicx}  

\usepackage{hyperref}
\usepackage{amsmath}
\usepackage{multirow}
\usepackage{amsfonts}
\usepackage{bm}
\usepackage{booktabs}
\usepackage{microtype}
\begin{document}
%
\title{Character $n$-gram Embeddings to Improve RNN Language Models}
\author{Sho Takase$^\dagger$\thanks{Current affiliation: Tokyo Institute of Technology.} \hspace{1.5em} Jun Suzuki$^{\dagger \ddagger}$ \hspace{1.5em} Masaaki Nagata$^\dagger$ \\
  $\dagger$NTT Communication Science Laboratories \\ $\ddagger$Tohoku University \\
  { sho.takase@nlp.c.titech.ac.jp} \hspace{1.5em} { jun.suzuki@ecei.tohoku.ac.jp} \hspace{1.5em} { nagata.masaaki@lab.ntt.co.jp} 
  }
\maketitle
\begin{abstract}
  This paper proposes a novel Recurrent Neural Network (RNN) language model that takes advantage of character information.
  We focus on character $n$-grams based on research in the field of word embedding construction~\cite{wieting-EtAl:2016:EMNLP2016}.
  Our proposed method constructs word embeddings from character $n$-gram embeddings and combines them with ordinary word embeddings.
  We demonstrate that the proposed method achieves the best perplexities on the language modeling datasets: Penn Treebank, WikiText-2, and WikiText-103.
  Moreover, we conduct experiments on application tasks: machine translation and headline generation.
  The experimental results indicate that our proposed method also positively affects these tasks.
\end{abstract}

\section{Introduction}\label{sec:intro}
Neural language models have played a crucial role in recent advances of neural network based methods in natural language processing (NLP).
For example, neural encoder-decoder models, which are becoming the de facto standard for various natural language generation tasks including machine translation~\cite{Sutskever:2014:SSL:2969033.2969173}, summarization~\cite{rush-chopra-weston:2015:EMNLP}, dialogue~\cite{wen-EtAl:2015:EMNLP}, and caption generation~\cite{Vinyals_2015_CVPR} can be interpreted as conditional neural language models.
Moreover, neural language models can be used for rescoring outputs from traditional methods, and they significantly improve the performance of automatic speech recognition~\cite{chime4asr}.
This implies that better neural language models improve the performance of application tasks.

In general, neural language models require word embeddings as an input~\cite{DBLP:journals/corr/ZarembaSV14}.
However, as described by \cite{verwimp-EtAl:2017:EACLlong}, this approach cannot make use of the internal structure
of words although the internal structure is often an effective clue for considering the meaning of a word.
For example, we can comprehend that the word `causal' is related to `cause' immediately because both words include the same character sequence `caus'.
Thus, if we incorporate a method that handles the internal structure such as character information, we can improve the quality of neural language models and probably make them robust to infrequent words.

To incorporate the internal structure, \cite{verwimp-EtAl:2017:EACLlong} concatenated character embeddings with an input word embedding.
They demonstrated that incorporating character embeddings improved the performance of RNN language models.
Moreover, \cite{Kim:2016:CNL:3016100.3016285} and \cite{45446} applied Convolutional Neural Networks (CNN) to construct word embeddings from character embeddings.

On the other hand, in the field of word embedding construction, some previous researchers found that character $n$-grams are more useful than single characters~\cite{wieting-EtAl:2016:EMNLP2016,TACL999}.
In particular, \cite{wieting-EtAl:2016:EMNLP2016} demonstrated that constructing word embeddings from character $n$-gram embeddings outperformed the methods that construct word embeddings from character embeddings by using CNN or a Long Short-Term Memory (LSTM).

Based on their reports, in this paper, we propose a neural language model that utilizes character $n$-gram embeddings.
Our proposed method encodes character $n$-gram embeddings to a word embedding with simplified Multi-dimensional Self-attention (MS)~\cite{shen2018disan}.
We refer to this constructed embedding as char$n$-MS-vec.
The proposed method regards char$n$-MS-vec as an input in addition to a word embedding.

We conduct experiments on the well-known benchmark datasets: Penn Treebank, WikiText-2, and WikiText-103.
Our experiments indicate that the proposed method outperforms neural language models trained with well-tuned hyperparameters and achieves state-of-the-art scores on each dataset.
In addition, we incorporate our proposed method into a standard neural encoder-decoder model and investigate its effect on machine translation and headline generation.
We indicate that the proposed method also has a positive effect on such tasks.

\section{RNN Language Model}
In this study, we focus on RNN language models, which are widely used in the literature.
This section briefly overviews the basic RNN language model.

In language modeling, we compute joint probability by using the product of conditional probabilities.
Let $w_{1:T}$ be a word sequence with length $T$, namely, $w_1, ..., w_T$.
We formally obtain the joint probability of word sequence $w_{1:T}$ as follows:
\begin{align}
  p(w_{1:T}) = p(w_1)\prod_{t=1}^{T-1} p(w_{t+1} | w_{1:t}). \label{eq:def_lm}
\end{align}
$p(w_1)$ is generally assumed to be $1$ in this literature, i.e., $p(w_1)\!=\!1$, and thus we can ignore its calculation\footnote{This definition is based on published implementations of previous studies such as https://github.com/wojzaremba/lstm and https://github.com/salesforce/awd-lstm-lm.}.

To estimate the conditional probability $p(w_{t+1} | w_{1:t})$, RNN language models encode sequence $w_{1:t}$ into a fixed-length vector and compute the probability distribution of each word from this fixed-length vector.
Let $V$ be the vocabulary size and let $P_{t} \in \mathbb{R}^{V}$ be the probability distribution of the vocabulary at timestep $t$.
Moreover, let $D_{h}$ be the dimension of the hidden state of an RNN and let $D_e$ be the dimensions of embedding vectors.
Then, RNN language models predict the probability distribution $P_{t+1}$ by the following equation:
\begin{align}
  P_{t+1} &= {\rm softmax}(W h_t + b), \label{eq:softmax} \\
  h_t &= f(e_t, h_{t-1}), \label{eq:rnn} \\
  e_t &= E x_t, \label{eq:embed}
\end{align}
where $W \in \mathbb{R}^{V \times D_h}$ is a weight matrix, $b \in \mathbb{R}^{V}$ is a bias term, and $E \in \mathbb{R}^{D_e \times V}$ is a word embedding matrix.
$x_t \in \{0,1\}^{V}$ and $h_{t} \in \mathbb{R}^{D_h}$ are a one-hot vector of an input word $w_t$ and the hidden state of the RNN at timestep $t$, respectively.
We define $h_{t}$ at timestep $t=0$ as a zero vector, that is, $h_0 = \bm{0}$.
Let $f(\cdot)$ represent an abstract function of an RNN, which might be the LSTM, the Quasi-Recurrent Neural Network (QRNN)~\cite{DBLP:journals/corr/BradburyMXS16}, or any other RNN variants.

\section{Incorporating Character \texorpdfstring{$n$}{n}-gram Embeddings}

\begin{figure}[!t]
  \centering
  \includegraphics[width=7cm]{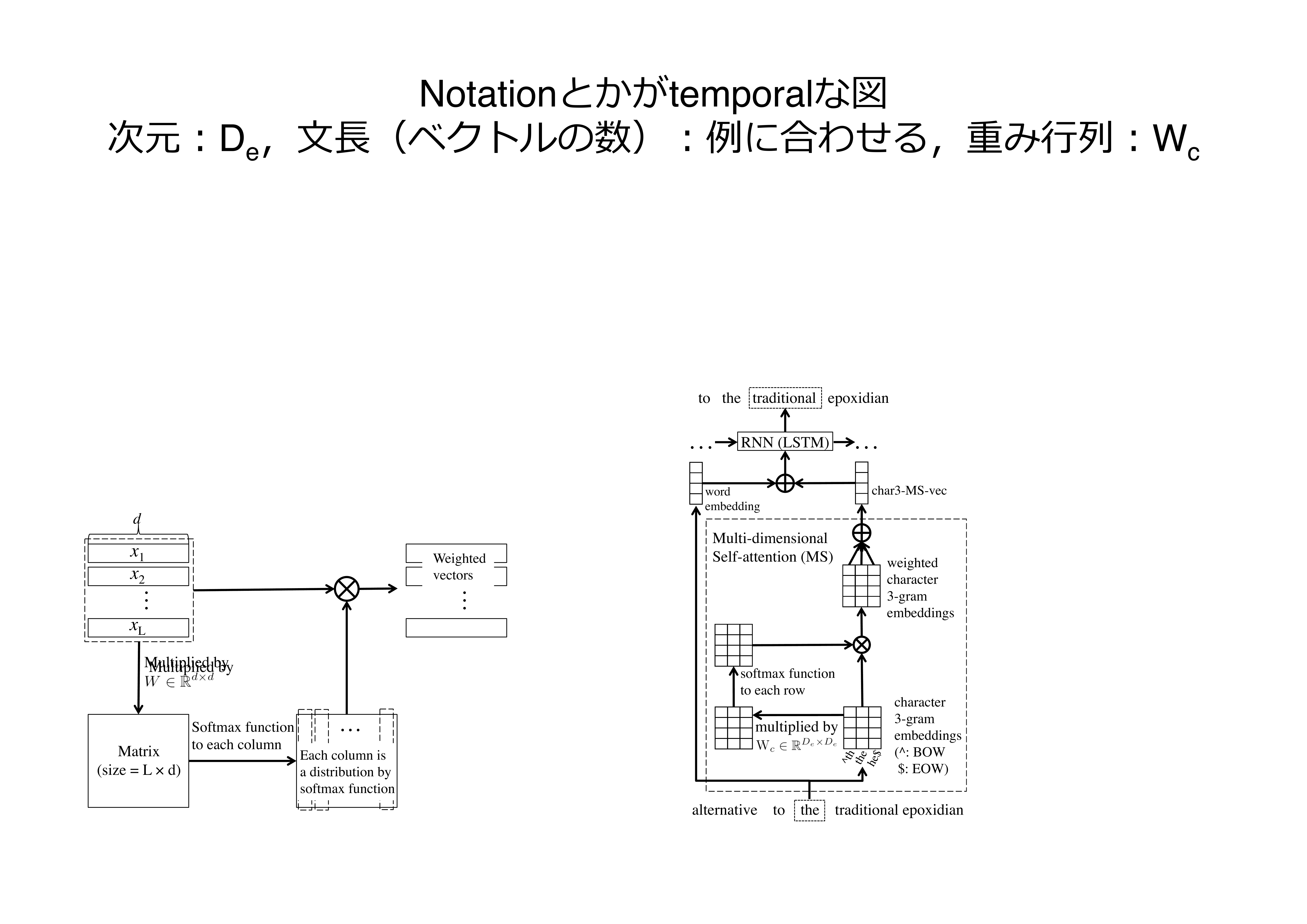}
   \caption{Overview of the proposed method. The proposed method computes char$n$-MS-vec from character $n$-gram (3-gram in this figure) embeddings and inputs the sum of it and the standard word embedding into an RNN.}
   \label{fig:overview}
\end{figure}

We incorporate char$n$-MS-vec, which is an embedding constructed from character $n$-gram embeddings, into RNN language models since, as discussed earlier, previous studies revealed that we can construct better word embeddings by using character $n$-gram embeddings~\cite{wieting-EtAl:2016:EMNLP2016,TACL999}.
In particular, we expect char$n$-MS-vec to help represent infrequent words by taking advantage of the internal structure.

Figure~\ref{fig:overview} is the overview of the proposed method using character 3-gram embeddings (char3-MS-vec).
As illustrated in this figure, our proposed method regards the sum of char3-MS-vec and the standard word embedding as an input of an RNN.
In other words, let $c_t$ be char$n$-MS-vec and we replace Equation \ref{eq:embed} with the following:
\begin{align}
  e_t &= E x_t + c_t. \label{eq:embed4proposed}
\end{align}

\subsection{Multi-dimensional Self-attention}

To compute $c_t$, we apply an encoder to character $n$-gram embeddings.
Previous studies demonstrated that additive composition, which computes the (weighted) sum of embeddings, is a suitable method for embedding construction~\cite{takase-okazaki-inui:2016:P16-1,wieting-EtAl:2016:EMNLP2016}.
Thus, we adopt (simplified) multi-dimensional self-attention~\cite{shen2018disan}, which computes weights for each dimension of given embeddings and sums up the weighted embeddings (i.e., element-wise weighted sum) as an encoder.
Let $s_i$ be the character $n$-gram embeddings of an input word, let $I$ be the number of character $n$-grams extracted from the word, and let $S$ be the matrix whose $i$-th column corresponds to $s_i$, that is, $S = \left[ s_1, ..., s_I \right]$.
The multi-dimensional self-attention constructs the word embedding $c_t$ by the following equations:
\begin{align}
  c_t &= \sum_{i=1}^{I} g_i \odot s_i, \label{eq:weightedsum} \\
  \{ g_i \}_j &= \{ {\rm softmax}( \left[ (W_c S)^{\mathsf{T}} \right]_{j} ) \}_i, \label{eq:multi_attention}
\end{align}
where $\odot$ means element-wise product of vectors, $W_c \in \mathbb{R}^{D_e \times D_e}$ is a weight matrix, $\left[ \cdot \right]_{j}$ is the $j$-th column of a given matrix, and $\{ \cdot \}_j$ is the $j$-th element of a given vector.
In short, Equation \ref{eq:multi_attention} applies the softmax function to each row of $\left[W_c S \right]$ and extracts the $i$-th column as $g_i$.

Let us consider the case where an input word is `the' and we use character 3-gram in Figure \ref{fig:overview}.
We prepare special characters `\^{}' and `\$' to represent the beginning and end of the word, respectively.
Then, `the' is composed of three character 3-grams: `\^{}th', `the', and `he\$'.
We multiply the embeddings of these 3-grams by transformation matrix $W_c$ and apply the softmax function to each row\footnote{\cite{shen2018disan} applied an activation function and one more transformation matrix to embeddings but we omit these operations.} as in Equation \ref{eq:multi_attention}.
As a result of the softmax, we obtain a matrix that contains weights for each embedding.
The size of the computed matrix is identical to the input embedding matrix: $D_e \times I$.
We then compute Equation \ref{eq:weightedsum}, i.e., the weighted sum of the embeddings, and add the resulting vector to the word embedding of `the'.
Finally, we input the vector into an RNN to predict the next word.

\subsection{Word Tying}
\cite{DBLP:journals/corr/InanKS16} and \cite{press-wolf:2017:EACLshort} proposed a word tying method (WT) that shares the word embedding matrix ($E$ in Equation \ref{eq:embed}) with the weight matrix to compute probability distributions ($W$ in Equation \ref{eq:softmax}).
They demonstrated that WT significantly improves the performance of RNN language models.

In this study, we adopt char$n$-MS-vec as the weight matrix in language modeling.
Concretely, we use $E + C$ instead of $W$ in Equation \ref{eq:softmax}, where $C \in \mathbb{R}^{D_e \times V}$ contains char$n$-MS-vec for all words in the vocabulary.

\section{Experiments on Language Modeling}

We investigate the effect of char$n$-MS-vec on the word-level language modeling task.
In detail, we examine the following four research questions;
\begin{enumerate}
  \item Can character $n$-gram embeddings improve the performance of state-of-the-art RNN language models?
  \item Do character $n$-gram embeddings have a positive effect on infrequent words?
  \item Is multi-dimensional self-attention effective for word embedding construction as compared with several other similar conventional methods?
  \item How many $n$ should we use?
\end{enumerate}

\begin{table}[!t]
  \centering
  \small
  \begin{tabular}{ c c | r | r | r }
  \toprule
  \multicolumn{2}{c|}{} & \multicolumn{1}{c|}{PTB} & \multicolumn{1}{c|}{WT2} & \multicolumn{1}{c}{WT103} \\
  \midrule
  \multicolumn{2}{c|}{Vocab} & 10,000 & 33,278 & 267,735 \\
  \midrule
   & Train & 929,590 & 2,088,628 & 103,227,021 \\
   \#Token & Valid & 73,761 & 217,646 & 217,646 \\
   & Test & 82,431 & 245,569 & 245,569 \\
  \bottomrule
  \end{tabular}
  \caption{Statistics of PTB, WT2, and WT103.\label{tab:dataset}}
\end{table}

\begin{table*}[!t]
  \centering
  \tabcolsep = 1.4mm
  \begin{tabular}{ l | c c r  r | c c r r | c c r r }
  \toprule
   & \multicolumn{4}{c}{PTB} & \multicolumn{4}{|c}{WT2} & \multicolumn{4}{|c}{WT103} \\
  Method & \#Char$n$ & \#Params & Valid & Test & \#Char$n$ & \#Params & Valid & Test & \#Char$n$ & \#Params & Valid & Test \\
  \midrule
  Baseline & - & 24M & 58.88 & 56.36 & - & 33M & 66.98 & 64.11 & - & 153M & 31.24 & 32.19 \\
  2 embeds & - & 28M & 58.94 & 56.56 & - & 47M & 66.71 & 64.00 & - & 261M & 31.75 & 32.71 \\
  \midrule
  char2-MS-vec & 773 & 25M & 57.88 & 55.91 & 2,442 & 35M & 65.60 & 62.96 & 8,124 & 158M & 31.18 & 31.92 \\
  char3-MS-vec & 5,258 & 26M & 57.40 & {\bf 55.56} & 12,932 & 39M & 65.05 & {\bf 61.95} & 51,492 & 175M & {\bf 30.95} & {\bf 31.81} \\
  char4-MS-vec & 15,416 & 31M & {\bf 57.30} & 55.64 & 39,318 & 49M & {\bf 64.84} & 62.19 & 179,900 & 226M & 31.23 & 32.21 \\
  \bottomrule
  \end{tabular}
  \caption{Perplexities on each dataset. We varied the $n$ for char$n$-MS-vec from 2 to 4.\label{tab:perplexities4charn}}
\end{table*}

\begin{table}[!t]
  \centering
  \begin{tabular}{ l | c  }
  \toprule
  Method & Seconds / epoch \\
  \midrule
  Baseline & \ \ 63.78 \\
  char2-MS-vec & 173.77 \\
  char3-MS-vec & 171.10 \\
  char4-MS-vec & 165.46 \\
  \bottomrule
  \end{tabular}
  \caption{Computational speed of the baseline and proposed method on NVIDIA Tesla P100.\label{tab:calc_speed}}
\end{table}

\begin{table}[!t]
  \centering
  \begin{tabular}{ l | c  c  }
  \toprule
  Method & Valid & Test \\
  \midrule
  Baseline & 44.02 & 41.14 \\
  \midrule
  char2-MS-vec & 43.21 & 40.99 \\
  char3-MS-vec & 42.74 & 40.50 \\
  char4-MS-vec & 42.71 & 40.41 \\
  \bottomrule
  \end{tabular}
  \caption{Perplexities on the PTB dataset where an input word is infrequent in the training data, which means its frequency is lower than 2,000.\label{tab:perplexities4freq}}
\end{table}

\subsection{Datasets}
We used the standard benchmark datasets for the word-level language modeling: Penn Treebank (PTB)~\cite{Marcus:1993:BLA:972470.972475}, WikiText-2 (WT2), and WikiText-103 (WT103)~\cite{DBLP:journals/corr/MerityXBS16}.
\cite{DBLP:conf/interspeech/MikolovKBCK10} and \cite{DBLP:journals/corr/MerityXBS16} published pre-processed PTB\footnote{http://www.fit.vutbr.cz/\~{}mikolov/rnnlm/}, WT2, and WT103\footnote{https://einstein.ai/research/the-wikitext-long-term-dependency-language-modeling-dataset}.
Following the previous studies, we used these pre-processed datasets for our experiments.

Table \ref{tab:dataset} describes the statistics of the datasets.
Table \ref{tab:dataset} demonstrates that the vocabulary size of WT103 is too large, and thus it is impractical to compute char$n$-MS-vec for all words at every step.
Therefore, we did not use $C$ for word tying.
In other words, we used only word embeddings $E$ as the weight matrix $W$ in WT103.

\subsection{Baseline RNN Language Model}
For base RNN language models, we adopted the state-of-the-art LSTM language model~\cite{merityRegOpt} for PTB and WT2, and QRNN for WT103~\cite{DBLP:journals/corr/BradburyMXS16}.
\cite{DBLP:journals/corr/MelisDB17} demonstrated that the standard LSTM trained with appropriate hyperparameters outperformed various architectures such as Recurrent Highway Networks (RHN)~\cite{zilly2016recurrent}.
In addition to several regularizations, \cite{merityRegOpt} introduced Averaged Stochastic Gradient Descent (ASGD)~\cite{polyak1992acceleration} to train the 3-layered LSTM language model.
As a result, their ASGD Weight-Dropped LSTM (AWD-LSTM) achieved state-of-the-art results on PTB and WT2.
For WT103, \cite{merityAnalysis} achieved the top score with the 4-layered QRNN.
Thus, we used AWD-LSTM for PTB and WT2, and QRNN for WT103 as the base language models, respectively.
We used their implementations\footnote{https://github.com/salesforce/awd-lstm-lm} for our experiments.

\subsection{Results}

Table \ref{tab:perplexities4charn} shows perplexities of the baselines and the proposed method.
We varied $n$ for char$n$-MS-vec from 2 to 4.
For the baseline, we also applied two word embeddings to investigate the performance in the case where we use more kinds of word embeddings.
In detail, we prepared $E_1, E_2 \in \mathbb{R}^{D_e \times V}$ and used $E_1 + E_2$ instead of $E$ in Equation \ref{eq:embed}.
Table \ref{tab:perplexities4charn} also shows the number of character $n$-grams in each dataset.
This table indicates that char$n$-MS-vec improved the performance of state-of-the-art models except for char4-MS-vec on WT103.
These results indicate that char$n$-MS-vec can raise the quality of word-level language models.
In particular, Table \ref{tab:perplexities4charn} shows that char3-MS-vec achieved the best scores consistently.
In contrast, an additional word embedding did not improve the performance.
This fact implies that the improvement of char$n$-MS-vec is caused by using character $n$-grams.
Thus, we answer yes to the first research question.

\begin{table}[!t]
  \centering
  \small
  \tabcolsep = 2mm
  \begin{tabular}{ l | c | c  c  }
  \toprule
  Method & \#Params & Valid & Test \\
  \midrule
  Baseline  & 24M & 58.88 & 56.36 \\
  \midrule
  \multicolumn{4}{c}{The proposed method (char$n$-MS-vec)} \\
  \midrule
  char2-MS-vec & 25M & 57.88 & 55.91 \\
  char3-MS-vec & 26M & 57.40 & {\bf 55.56} \\
  char4-MS-vec & 31M & {\bf 57.30} & 55.64 \\
  \midrule
  \multicolumn{4}{c}{Using CNN (charCNN)} \\
  \midrule
  Original Settings & 36M & 60.59 & 58.03 \\
  Small CNN result dims & 21M & 82.39 & 80.01\\
  Large embedding dims & 36M & 60.75 & 58.06 \\
  \midrule
  \multicolumn{4}{c}{Using Sum (char$n$-Sum-vec)} \\
  \midrule
  char2-Sum-vec & 25M & 61.77 & 59.57 \\
  char3-Sum-vec & 26M & 59.72 & 57.39 \\
  char4-Sum-vec & 30M & 59.40 & 56.72 \\
  \midrule
  \multicolumn{4}{c}{Using Standard self-attention (char$n$-SS-vec)} \\
  \midrule
  char2-SS-vec & 25M & 58.66 & 56.44 \\
  char3-SS-vec & 26M & 59.42 & 57.30 \\
  char4-SS-vec & 30M & 58.37 & 56.33 \\
  \midrule
  \multicolumn{4}{c}{Exclude $C$ from word tying} \\
  \midrule
  char2-MS-vec & 25M & 58.10 & 56.12 \\
  char3-MS-vec & 26M & 58.62 & 56.19 \\
  char4-MS-vec & 31M & 58.68 & 56.78 \\
  \midrule
  \multicolumn{4}{c}{Remove word embeddings $E$} \\
  \midrule
  char2-MS-vec & 21M & 74.22 & 72.39 \\
  char3-MS-vec & 22M & 60.60 & 58.88 \\
  char4-MS-vec & 27M & 57.65 & 55.74 \\
  \midrule
  \multicolumn{4}{c}{Same \#Params as baseline} \\
  \midrule
  char2-MS-vec & 24M & 57.93 & 56.15 \\
  char3-MS-vec & 24M & 57.81 & 55.81 \\
  char4-MS-vec & 24M & 59.48 & 57.69 \\  
  \bottomrule
  \end{tabular}
  \caption{Perplexities of each structure on PTB dataset.\label{tab:ablation}}
\end{table}

\begin{table*}[!t]
  \centering
  \begin{tabular}{ l  | c | c c }
  \toprule
  Model & \#Params & Valid & Test \\
  \midrule
  LSTM (medium) \cite{DBLP:journals/corr/ZarembaSV14} & 20M & 86.2 \ \  & 82.7 \ \  \\
  LSTM (large) \cite{DBLP:journals/corr/ZarembaSV14} & 66M & 82.2 \ \  & 78.4 \ \  \\
  Character-Word LSTM \cite{verwimp-EtAl:2017:EACLlong} & - & - & 82.04 \ \  \\
  LSTM-CharCNN-Large \cite{Kim:2016:CNL:3016100.3016285} & 19M & - & 78.9 \ \  \\
  Variational LSTM (medium) \cite{Gal2016Theoretically} & 20M & 81.9 $\pm$ 0.2 & 79.7 $\pm$ 0.1 \\
  Variational LSTM (large) \cite{Gal2016Theoretically} & 66M & 77.9 $\pm$ 0.3 & 75.2 $\pm$ 0.2 \\
  Variational RHN \cite{zilly2016recurrent} & 32M & 71.2 \ \  & 68.5 \ \  \\
  Variational RHN + WT \cite{zilly2016recurrent} & 23M & 67.9 \ \  & 65.4 \ \  \\
  Variational RHN + WT + IOG \cite{takase-suzuki-nagata:2017:I17-2} & 29M & 67.0 \ \  & 64.4 \ \  \\
  Neural Architecture Search \cite{45826} & 54M & - & 62.4 \ \  \\
  LSTM with skip connections \cite{DBLP:journals/corr/MelisDB17} & 24M & 60.9 \ \  & 58.3 \ \  \\
  AWD-LSTM \cite{merityRegOpt} & 24M & 60.0 \ \  & 57.3 \ \  \\
  AWD-LSTM + Fraternal Dropout \cite{fraternal} & 24M & 58.9 \ \  & 56.8 \ \  \\
  AWD-LSTM-MoS \cite{DBLP:journals/corr/abs-1711-03953} & 22M & 56.54 & 54.44 \\
  AWD-LSTM-DOC \cite{D18-1489} & 23M & 54.12 & 52.38 \\  
  \midrule
  Proposed method: AWD-LSTM + char3-MS-vec & 26M & {\bf 57.40} & {\bf 55.56} \\
  Proposed method: AWD-LSTM-MoS + char3-MS-vec & 23M & {\bf 56.13} & {\bf 53.98} \\
  Proposed method: AWD-LSTM-DOC + char3-MS-vec & 24M & {\bf 53.76} & {\bf 52.34} \\
  \bottomrule
  \end{tabular}
  \caption{Perplexities of the proposed method and as reported in previous studies on the PTB dataset.\label{tb:perplexity}}
\end{table*}

\begin{table*}[!t]
  \centering
  \begin{tabular}{ l | c | c c }
  \toprule
  Model & \#Params & Valid & Test \\
  \midrule
  LSTM \cite{DBLP:journals/corr/GraveJU16} & - & - & 99.3 \ \  \\
  Variational LSTM + IOG \cite{takase-suzuki-nagata:2017:I17-2} & 70M & 95.9 \ \  & 91.0 \ \  \\
  Variational LSTM + WT + AL \cite{DBLP:journals/corr/InanKS16} & 28M & 91.5 \ \  & 87.0 \ \  \\
  LSTM with skip connections \cite{DBLP:journals/corr/MelisDB17} & 24M & 69.1 \ \  & 65.9 \ \  \\
  AWD-LSTM \cite{merityRegOpt} & 33M & 68.6 \ \  & 65.8 \ \  \\
  AWD-LSTM + Fraternal Dropout \cite{fraternal} & 34M & 66.8 \ \  & 64.1 \ \  \\
  AWD-LSTM-MoS \cite{DBLP:journals/corr/abs-1711-03953} & 35M & 63.88 & 61.45 \\
  AWD-LSTM-DOC \cite{D18-1489} & 37M & 60.29 & 58.03 \\
  \midrule
  Proposed method: AWD-LSTM + char3-MS-vec & 39M & {\bf 65.05} & {\bf 61.95} \\
  Proposed method: AWD-LSTM-MoS + char3-MS-vec & 39M & {\bf 62.20} & {\bf 59.99} \\
  Proposed method: AWD-LSTM-DOC + char3-MS-vec & 41M & {\bf 59.47} & {\bf 57.30} \\
  \bottomrule
  \end{tabular}
  \caption{Perplexities of the proposed method and as reported in previous studies on the WT2 dataset.\label{tb:perplexityOnWikitext2}}
\end{table*}

\begin{table*}[!t]
  \centering
  \begin{tabular}{ l | c | c c }
  \toprule
  Model & \#Params & Valid & Test \\ 
  \midrule
  LSTM \cite{DBLP:journals/corr/GraveJU16} & - & - & 48.7 \\
  GCNN-8 \cite{DBLP:journals/corr/DauphinFAG16} & - & - & 44.9 \\
  GCNN-14 \cite{DBLP:journals/corr/DauphinFAG16} & - & - & 37.2 \\
  QRNN \cite{merityAnalysis} & 153M & 32.0 & 33.0 \\
  \midrule
  Proposed method: QRNN + char3-MS-vec & 175M & {\bf 30.95} & {\bf 31.81} \\
  \bottomrule
  \end{tabular}
  \caption{Perplexities of the proposed method and as reported in previous studies on the WT103 dataset.\label{tb:perplexityOnWikitext103}}
\end{table*}

Table \ref{tab:calc_speed} shows the training time spent on each epoch.
We calculated it on the NVIDIA Tesla P100.
Table \ref{tab:calc_speed} indicates that the proposed method requires more computational time than the baseline unfortunately.
We leave exploring a faster structure for our future work.

Table \ref{tab:perplexities4freq} shows perplexities on the PTB dataset where the frequency of an input word is lower than 2,000 in the training data.
This table indicates that the proposed method can improve the performance even if an input word is infrequent.
In other words, char$n$-MS-vec helps represent the meanings of infrequent words.
Therefore, we answer yes to the second research question in the case of our experimental settings.

We explored the effectiveness of multi-dimensional self-attention for word embedding construction.
Table \ref{tab:ablation} shows perplexities of using several encoders on the PTB dataset.
As in \cite{Kim:2016:CNL:3016100.3016285}, we applied CNN to construct word embeddings (charCNN in Table \ref{tab:ablation}).
Moreover, we applied the summation and standard self-attention, which computes the scalar value as a weight for a character $n$-gram embedding, to construct word embeddings (char$n$-Sum-vec and char$n$-SS-vec, respectively).
For CNN, we used hyperparameters identical to \cite{Kim:2016:CNL:3016100.3016285} (``Original Settings'' in Table \ref{tab:ablation}) but the setting has two differences from other architectures: 1. The dimension of the computed vectors is much larger than the dimension of the baseline word embeddings and 2. The dimension of the input character embeddings is much smaller than the dimension of the baseline word embeddings.
Therefore, we added two configurations: assigning the dimension of the computed vectors and input character embeddings a value identical to the baseline word embeddings (in Table \ref{tab:ablation}, ``Small CNN result dims'' and ``Large embedding dims'', respectively).

Table \ref{tab:ablation} shows that the proposed char$n$-MS-vec outperformed charCNN even though the original settings of charCNN had much larger parameters.
Moreover, we trained charCNN with two additional settings but CNN did not improve the baseline performance.
This result implies that char$n$-MS-vec is better embeddings than ones constructed by applying CNN to character embeddings.
Table \ref{tab:ablation} also indicates that char$n$-Sum-vec was harmful to the performance.
Moreover, char$n$-SS-vec did not have a positive effect on the baseline.
These results answer yes to the third research question; our use of multi-dimensional self-attention is more appropriate for constructing word embeddings from character $n$-gram embeddings.

Table \ref{tab:ablation} also shows that excluding $C$ from word tying (``Exclude $C$ from word tying'') achieved almost the same score as the baseline.
Moreover, this table indicates that performance fails as the the number of parameters is increased.
Thus, we need to assign $C$ to word tying to prevent over-fitting for the PTB dataset.
In addition, this result implies that the performance of WT103 in Table \ref{tab:perplexities4charn} might be raised if we can apply word tying to WT103.

Moreover, to investigate the effect of only char$n$-MS-vec, we ignore $E x_t$ in Equation \ref{eq:embed4proposed}.
We refer to this setting as ``Remove word embeddings $E$'' in Table \ref{tab:ablation}.
Table \ref{tab:ablation} shows cahr3-MS-vec and char4-MS-vec are superior to char2-MS-vec.
In the view of perplexity, char3-MS-vec and char4-MS-vec achieved comparable scores to each other.
On the other hand, char3-MS-vec is composed of much smaller parameters.
Furthermore, we decreased the embedding size $D_e$ to adjust the number of parameters to the same size as the baseline (``Same \#Params as baseline'' in Table \ref{tab:ablation}).
In this setting, char3-MS-vec achieved the best perplexity.
Therefore, we consider that char3-MS-vec is more useful than char4-MS-vec, which is the answer to the fourth research question.
We use the combination of the char3-MS-vec $c_t$ and word embedding $E x_t$ in the following experiments.

Finally, we compare the proposed method with the published scores reported in previous studies.
Tables \ref{tb:perplexity}, \ref{tb:perplexityOnWikitext2}, and \ref{tb:perplexityOnWikitext103}, respectively, show perplexities of the proposed method and previous studies on PTB, WT2, and WT103\footnote{To compare models trained only on the training data, we excluded methods that use the statistics of the test data~\cite{DBLP:journals/corr/GraveJU16,DBLP:journals/corr/abs-1709-07432}.}.
Since AWD-LSTM-MoS~\cite{DBLP:journals/corr/abs-1711-03953} and AWD-LSTM-DOC~\cite{D18-1489} achieved the state-of-the-art scores on PTB and WT2, we combined char3-MS-vec with them.
These tables show that the proposed method improved the performance of the base model and outperformed the state-of-the-art scores on all datasets.
In particular, char3-MS-vec improved perplexity by at least 1 point from current best scores on the WT103 dataset.

\section{Experiments on Applications}

As described in Section \ref{sec:intro}, neural encoder-decoder models can be interpreted as conditional neural language models.
Therefore, to investigate if the proposed method contributes to encoder-decoder models, we conduct experiments on machine translation and headline generation tasks.

\subsection{Datasets}
For machine translation, we used two kinds of language pairs: English-French and English-German sentences in the IWSLT 2016 dataset\footnote{https://wit3.fbk.eu/}.
The dataset contains about 208K English-French pairs and 189K English-German pairs.
We conducted four translation tasks: from English to each language (En-Fr and En-De), and their reverses (Fr-En and De-En).

\begin{table*}[!t]
  \centering
  \begin{tabular}{ l | c  c  c  c }
  \toprule
  Model & En-Fr & En-De & Fr-En & De-En \\
  \midrule
  EncDec & 34.37 & 23.05 & 34.07 & 28.18 \\
  EncDec+char3-MS-vec & {\bf 35.48} & {\bf 23.27} & {\bf 34.43} & {\bf 28.86} \\
  \bottomrule
  \end{tabular}
  \caption{BLEU scores on the IWSLT16 dataset. We report the average score of 3 runs.\label{tb:nmt}}
\end{table*}

\begin{table*}[!t]
  \centering
  \small
  \begin{tabular}{ l | c c c | c c c }
  \toprule
  & \multicolumn{3}{c|}{Test set by \cite{zhou-EtAl:2017:Long}} & \multicolumn{3}{c}{Test set by \cite{kiyono}} \\
  Model & ROUGE-1 & ROUGE-2 & ROUGE-L & ROUGE-1 & ROUGE-2 & ROUGE-L \\
  \midrule
  EncDec & 46.77 & 24.87 & 43.58 & 46.78 & 24.52 & 43.68 \\
  EncDec+char3-MS-vec & {\bf 47.04} & {\bf 25.09} & {\bf 43.80} & {\bf 46.91} & {\bf 24.61} & {\bf 43.77} \\
  \midrule
  ABS \cite{rush-chopra-weston:2015:EMNLP} & 37.41 & 15.87 & 34.70 & - & - & - \\
  SEASS \cite{zhou-EtAl:2017:Long} & 46.86 & 24.58 & 43.53 & - & - & - \\ 
  \cite{kiyono} & 46.34 & 24.85 & 43.49  & 46.41 & 24.58 & 43.59 \\
  \bottomrule
  \end{tabular}
  \caption{ROUGE F1 scores on the headline generation test sets provided by \protect\cite{zhou-EtAl:2017:Long} and \protect\cite{kiyono}. The upper part is the results of our implementation and the lower part shows the scores reported in previous studies. In the upper part, we report the average score of 3 runs.\label{tb:headline}}
\end{table*}

For headline generation, we used sentence-headline pairs extracted from the annotated English Gigaword corpus~\cite{Napoles:2012:AG:2391200.2391218} in the same manner as \cite{rush-chopra-weston:2015:EMNLP}.
The training set contains about 3.8M sentence-headline pairs.
For evaluation, we exclude the test set constructed by \cite{rush-chopra-weston:2015:EMNLP} because it contains some invalid instances, as reported in \cite{zhou-EtAl:2017:Long}.
We instead used the test sets constructed by \cite{zhou-EtAl:2017:Long} and \cite{kiyono}.

\subsection{Experimental Settings}
We employed the neural encoder-decoder with attention mechanism described in \cite{kiyono} as the base model.
Its encoder consists of a 2-layer bidirectional LSTM and its decoder consists of a 2-layer LSTM with attention mechanism proposed by \cite{luong-pham-manning:2015:EMNLP}.
We refer to this neural encoder-decoder as EncDec.
To investigate the effect of the proposed method, we introduced char3-MS-vec into EncDec.
Here, we applied char3-MS-vec to both the encoder and decoder.
Moreover, we did not apply word tying technique to EncDec because it is default setting in the widely-used encoder-decoder implementation\footnote{http://opennmt.net/}.

We set the embedding size and dimension of the LSTM hidden state to 500 for machine translation and 400 for headline generation.
The mini-batch size is 64 for machine translation and 256 for headline generation.
For other hyperparameters, we followed the configurations described in \cite{kiyono}.
We constructed the vocabulary set by using Byte-Pair-Encoding\footnote{https://github.com/rsennrich/subword-nmt} (BPE)~\cite{sennrich-haddow-birch:2016:P16-11} because BPE is a currently widely-used technique for vocabulary construction.
We set the number of BPE merge operations to 16K for machine translation and 5K for headline generation.

\subsection{Results}

Tables \ref{tb:nmt} and \ref{tb:headline} show the results of machine translation and headline generation, respectively.
These tables show that EncDec+char3-MS-vec outperformed EncDec in all test data.
In other words, these results indicate that our proposed method also has a positive effect on the neural encoder-decoder model.
Moreover, it is noteworthy that char3-MS-vec improved the performance of EncDec even though the vocabulary set constructed by BPE contains subwords.
This implies that character $n$-gram embeddings improve the quality of not only word embeddings but also subword embeddings.

In addition to the results of our implementations, the lower portion of Table \ref{tb:headline} contains results reported in previous studies.
Table \ref{tb:headline} shows that EncDec+char3-MS-vec also outperformed the methods proposed in previous studies.
Therefore, EncDec+char3-MS-vec achieved the top scores in the test sets constructed by \cite{zhou-EtAl:2017:Long} and \cite{kiyono} even though it does not have a task-specific architecture such as the selective gate proposed by \cite{zhou-EtAl:2017:Long}.

In these experiments, we only applied char3-MS-vec to EncDec but \cite{C18-1052} indicated that combining multiple kinds of subword units can improve the performance.
We will investigate the effect of combining several character $n$-gram embeddings in future work.

\section{Related Work}
\paragraph{RNN Language Model}
\cite{DBLP:conf/interspeech/MikolovKBCK10} introduced RNN into language modeling to handle arbitrary-length sequences in computing conditional probability $p(w_{t+1} | w_{1:t})$.
They demonstrated that the RNN language model outperformed the Kneser-Ney smoothed 5-gram language model~\cite{Chen:1996:ESS:981863.981904}, which is a sophisticated $n$-gram language model.

\cite{DBLP:journals/corr/ZarembaSV14} drastically improved the performance of language modeling by applying LSTM and the dropout technique~\cite{Srivastava:2014:DSW:2627435.2670313}.
\cite{DBLP:journals/corr/ZarembaSV14} applied dropout to all the connections except for recurrent connections but \cite{Gal2016Theoretically} proposed variational inference based dropout to regularize recurrent connections.
\cite{DBLP:journals/corr/MelisDB17} demonstrated that the standard LSTM can achieve superior performance by selecting appropriate hyperparameters.
Finally, \cite{merityRegOpt} introduced DropConnect~\cite{wan2013regularization} and averaged SGD~\cite{polyak1992acceleration} into the LSTM language model and achieved state-of-the-art perplexities on PTB and WT2.
For WT103, \cite{merityAnalysis} found that QRNN~\cite{DBLP:journals/corr/BradburyMXS16}, which is a faster architecture than LSTM, achieved the best perplexity.
Our experimental results show that the proposed char$n$-MS-vec improved the performance of these state-of-the-art language models.

\cite{DBLP:journals/corr/abs-1711-03953} explained that the training of RNN language models can be interpreted as matrix factorization.
In addition, to raise an expressive power, they proposed Mixture of Softmaxes (MoS) that computes multiple probability distributions from a final RNN layer and combines them with a weighted average.
\cite{D18-1489} proposed Direct Output Connection (DOC) that is a generalization of MoS.
They used middle layers in addition to the final layer to compute probability distributions.
These methods (AWD-LSTM-MoS and AWD-LSTM-DOC) achieved the current state-of-the-art perplexities on PTB and WT2.
Our proposed method can also be combined with MoS and DOC.
In fact, Tables \ref{tb:perplexity} and \ref{tb:perplexityOnWikitext2} indicate that the proposed method further improved the performance of them.

\cite{Kim:2016:CNL:3016100.3016285} introduced character information into RNN language models.
They applied CNN to character embeddings for word embedding construction.
Their proposed method achieved perplexity competitive with the basic LSTM language model~\cite{DBLP:journals/corr/ZarembaSV14} even though its parameter size is small.
\cite{45446} also applied CNN to construct word embeddings from character embeddings.
They indicated that CNN also positively affected the LSTM language model in a huge corpus.
\cite{verwimp-EtAl:2017:EACLlong} proposed a method concatenating character embeddings with a word embedding to use character information.
In contrast to these methods, we used character $n$-gram embeddings to construct word embeddings.
To compare the proposed method to these methods, we combined the CNN proposed by \cite{Kim:2016:CNL:3016100.3016285} with the state-of-the-art LSTM language model (AWD-LSTM)~\cite{merityRegOpt}.
Our experimental results indicate that the proposed method outperformed the method using character embeddings (charCNN in Table \ref{tab:ablation}).

Some previous studies focused on boosting the performance of language models during testing~\cite{DBLP:journals/corr/GraveJU16,DBLP:journals/corr/abs-1709-07432}.
For example, \cite{DBLP:journals/corr/abs-1709-07432} proposed dynamic evaluation that updates model parameters based on the given correct sequence during evaluation.
Although these methods might further improve our proposed language model, we omitted these methods since it is unreasonable to obtain correct outputs in applications such as machine translation.

\paragraph{Embedding Construction}
Previous studies proposed various methods to construct word embeddings.
\cite{luong-socher-manning:2013:CoNLL-2013} applied Recursive Neural Networks to construct word embeddings from morphemic embeddings.
\cite{ling-EtAl:2015:EMNLP2} applied bidirectional LSTMs to character embeddings for word embedding construction.
On the other hand, \cite{TACL999} and \cite{wieting-EtAl:2016:EMNLP2016} focused on character $n$-gram.
They demonstrated that the sum of character $n$-gram embeddings outperformed ordinary word embeddings.
In addition, \cite{wieting-EtAl:2016:EMNLP2016} found that the sum of character $n$-gram embeddings also outperformed word embeddings constructed from character embeddings with CNN and LSTM.

As an encoder, previous studies argued that additive composition, which computes the (weighted) sum of embeddings, is a suitable method theoretically~\cite{rantian:additive} and empirically~\cite{Muraoka-EtAl:2014:PACLIC,takase-okazaki-inui:2016:P16-1}.
In this paper, we used multi-dimensional self-attention to construct word embeddings because it can be interpreted as an element-wise weighted sum.
Through experiments, we indicated that multi-dimensional self-attention is superior to the summation and standard self-attention as an encoder.

\section{Conclusion}
In this paper, we incorporated character information with RNN language models.
Based on the research in the field of word embedding construction~\cite{wieting-EtAl:2016:EMNLP2016}, we focused on character $n$-gram embeddings to construct word embeddings.
We used multi-dimensional self-attention~\cite{shen2018disan} to encode character $n$-gram embeddings.
Our proposed char$n$-MS-vec improved the performance of state-of-the-art RNN language models and achieved the best perplexities on Penn Treebank, WikiText-2, and WikiText-103.
Moreover, we investigated the effect of char$n$-MS-vec on application tasks, specifically, machine translation and headline generation.
Our experiments show that char$n$-MS-vec also improved the performance of a neural encoder-decoder on both tasks.

\section*{Acknowledgments}
This work was supported by JSPS KAKENHI Grant Number JP18K18119.
We would like to thank the anonymous reviewers for their helpful suggestions and comments.

\fontsize{9.0pt}{10.0pt}
\bibliography{AAAI-TakaseS.7094.bbl}

\begin{thebibliography}{}

\bibitem[\protect\citeauthoryear{Bojanowski \bgroup et al\mbox.\egroup
  }{2017}]{TACL999}
Bojanowski, P.; Grave, E.; Joulin, A.; and Mikolov, T.
\newblock 2017.
\newblock Enriching word vectors with subword information.
\newblock {\em TACL} 5:135--146.

\bibitem[\protect\citeauthoryear{Bradbury \bgroup et al\mbox.\egroup
  }{2017}]{DBLP:journals/corr/BradburyMXS16}
Bradbury, J.; Merity, S.; Xiong, C.; and Socher, R.
\newblock 2017.
\newblock Quasi-recurrent neural networks.
\newblock In {\em Proceedings of ICLR}.

\bibitem[\protect\citeauthoryear{Chen and
  Goodman}{1996}]{Chen:1996:ESS:981863.981904}
Chen, S.~F., and Goodman, J.
\newblock 1996.
\newblock An empirical study of smoothing techniques for language modeling.
\newblock In {\em Proceedings of ACL},  310--318.

\bibitem[\protect\citeauthoryear{Dauphin \bgroup et al\mbox.\egroup
  }{2016}]{DBLP:journals/corr/DauphinFAG16}
Dauphin, Y.~N.; Fan, A.; Auli, M.; and Grangier, D.
\newblock 2016.
\newblock Language modeling with gated convolutional networks.
\newblock {\em CoRR} abs/1612.08083.

\bibitem[\protect\citeauthoryear{Du \bgroup et al\mbox.\egroup
  }{2016}]{chime4asr}
Du, J.; Tu, Y.-H.; Sun, L.; Ma, F.; Wang, H.-K.; Pan, J.; Liu, C.; and Lee,
  C.-H.
\newblock 2016.
\newblock The ustc-iflytek system for chime-4 challenge.
\newblock In {\em Proceedings of CHiME-4},  36--38.

\bibitem[\protect\citeauthoryear{Gal and
  Ghahramani}{2016}]{Gal2016Theoretically}
Gal, Y., and Ghahramani, Z.
\newblock 2016.
\newblock {A Theoretically Grounded Application of Dropout in Recurrent Neural
  Networks}.
\newblock In {\em Proceedings of NIPS}.

\bibitem[\protect\citeauthoryear{Grave, Joulin, and
  Usunier}{2017}]{DBLP:journals/corr/GraveJU16}
Grave, E.; Joulin, A.; and Usunier, N.
\newblock 2017.
\newblock {Improving Neural Language Models with a Continuous Cache}.
\newblock In {\em Proceedings of ICLR}.

\bibitem[\protect\citeauthoryear{Inan, Khosravi, and
  Socher}{2017}]{DBLP:journals/corr/InanKS16}
Inan, H.; Khosravi, K.; and Socher, R.
\newblock 2017.
\newblock {Tying Word Vectors and Word Classifiers: {A} Loss Framework for
  Language Modeling}.
\newblock In {\em Proceedings of ICLR}.

\bibitem[\protect\citeauthoryear{Jozefowicz \bgroup et al\mbox.\egroup
  }{2016}]{45446}
Jozefowicz, R.; Vinyals, O.; Schuster, M.; Shazeer, N.; and Wu, Y.
\newblock 2016.
\newblock Exploring the limits of language modeling.

\bibitem[\protect\citeauthoryear{Kim \bgroup et al\mbox.\egroup
  }{2016}]{Kim:2016:CNL:3016100.3016285}
Kim, Y.; Jernite, Y.; Sontag, D.; and Rush, A.~M.
\newblock 2016.
\newblock Character-aware neural language models.
\newblock In {\em Proceedings of AAAI},  2741--2749.

\bibitem[\protect\citeauthoryear{Kiyono \bgroup et al\mbox.\egroup
  }{2017}]{kiyono}
Kiyono, S.; Takase, S.; Suzuki, J.; Okazaki, N.; Inui, K.; and Nagata, M.
\newblock 2017.
\newblock Source-side prediction for neural headline generation.
\newblock {\em CoRR}.

\bibitem[\protect\citeauthoryear{Krause \bgroup et al\mbox.\egroup
  }{2017}]{DBLP:journals/corr/abs-1709-07432}
Krause, B.; Kahembwe, E.; Murray, I.; and Renals, S.
\newblock 2017.
\newblock Dynamic evaluation of neural sequence models.
\newblock {\em CoRR}.

\bibitem[\protect\citeauthoryear{Ling \bgroup et al\mbox.\egroup
  }{2015}]{ling-EtAl:2015:EMNLP2}
Ling, W.; Dyer, C.; Black, A.~W.; Trancoso, I.; Fermandez, R.; Amir, S.;
  Marujo, L.; and Luis, T.
\newblock 2015.
\newblock Finding function in form: Compositional character models for open
  vocabulary word representation.
\newblock In {\em Proceedings of EMNLP},  1520--1530.

\bibitem[\protect\citeauthoryear{Luong, Pham, and
  Manning}{2015}]{luong-pham-manning:2015:EMNLP}
Luong, T.; Pham, H.; and Manning, C.~D.
\newblock 2015.
\newblock Effective approaches to attention-based neural machine translation.
\newblock In {\em Proceedings of EMNLP},  1412--1421.

\bibitem[\protect\citeauthoryear{Luong, Socher, and
  Manning}{2013}]{luong-socher-manning:2013:CoNLL-2013}
Luong, T.; Socher, R.; and Manning, C.
\newblock 2013.
\newblock Better word representations with recursive neural networks for
  morphology.
\newblock In {\em Proceedings of CoNLL},  104--113.

\bibitem[\protect\citeauthoryear{Marcus, Marcinkiewicz, and
  Santorini}{1993}]{Marcus:1993:BLA:972470.972475}
Marcus, M.~P.; Marcinkiewicz, M.~A.; and Santorini, B.
\newblock 1993.
\newblock {Building a Large Annotated Corpus of English: The Penn Treebank}.
\newblock {\em Computational Linguistics} 19(2):313--330.

\bibitem[\protect\citeauthoryear{Melis, Dyer, and
  Blunsom}{2018}]{DBLP:journals/corr/MelisDB17}
Melis, G.; Dyer, C.; and Blunsom, P.
\newblock 2018.
\newblock On the state of the art of evaluation in neural language models.
\newblock {\em Proceedings of ICLR}.

\bibitem[\protect\citeauthoryear{Merity \bgroup et al\mbox.\egroup
  }{2017}]{DBLP:journals/corr/MerityXBS16}
Merity, S.; Xiong, C.; Bradbury, J.; and Socher, R.
\newblock 2017.
\newblock {Pointer Sentinel Mixture Models}.
\newblock In {\em Proceedings of ICLR}.

\bibitem[\protect\citeauthoryear{Merity, Keskar, and
  Socher}{2018a}]{merityAnalysis}
Merity, S.; Keskar, N.~S.; and Socher, R.
\newblock 2018a.
\newblock {An Analysis of Neural Language Modeling at Multiple Scales}.
\newblock {\em arXiv preprint arXiv:1803.08240}.

\bibitem[\protect\citeauthoryear{Merity, Keskar, and
  Socher}{2018b}]{merityRegOpt}
Merity, S.; Keskar, N.~S.; and Socher, R.
\newblock 2018b.
\newblock {Regularizing and Optimizing LSTM Language Models}.
\newblock In {\em Proceedings of ICLR}.

\bibitem[\protect\citeauthoryear{Mikolov \bgroup et al\mbox.\egroup
  }{2010}]{DBLP:conf/interspeech/MikolovKBCK10}
Mikolov, T.; Karafi{\'{a}}t, M.; Burget, L.; Cernock{\'{y}}, J.; and Khudanpur,
  S.
\newblock 2010.
\newblock Recurrent neural network based language model.
\newblock In {\em Proceedings of INTERSPEECH},  1045--1048.

\bibitem[\protect\citeauthoryear{Morishita, Suzuki, and
  Nagata}{2018}]{C18-1052}
Morishita, M.; Suzuki, J.; and Nagata, M.
\newblock 2018.
\newblock Improving neural machine translation by incorporating hierarchical
  subword features.
\newblock In {\em Proceedings of COLING},  618--629.

\bibitem[\protect\citeauthoryear{Muraoka \bgroup et al\mbox.\egroup
  }{2014}]{Muraoka-EtAl:2014:PACLIC}
Muraoka, M.; Shimaoka, S.; Yamamoto, K.; Watanabe, Y.; Okazaki, N.; and Inui,
  K.
\newblock 2014.
\newblock Finding the best model among representative compositional models.
\newblock In {\em Proceedings of PACLIC},  65--74.

\bibitem[\protect\citeauthoryear{Napoles, Gormley, and
  Van~Durme}{2012}]{Napoles:2012:AG:2391200.2391218}
Napoles, C.; Gormley, M.; and Van~Durme, B.
\newblock 2012.
\newblock Annotated gigaword.
\newblock In {\em Proceedings of AKBC-WEKEX},  95--100.

\bibitem[\protect\citeauthoryear{Polyak and
  Juditsky}{1992}]{polyak1992acceleration}
Polyak, B.~T., and Juditsky, A.~B.
\newblock 1992.
\newblock {Acceleration of Stochastic Approximation by Averaging}.
\newblock {\em SIAM Journal on Control and Optimization} 30(4):838--855.

\bibitem[\protect\citeauthoryear{Press and
  Wolf}{2017}]{press-wolf:2017:EACLshort}
Press, O., and Wolf, L.
\newblock 2017.
\newblock {Using the Output Embedding to Improve Language Models}.
\newblock In {\em Proceedings of EACL},  157--163.

\bibitem[\protect\citeauthoryear{Rush, Chopra, and
  Weston}{2015}]{rush-chopra-weston:2015:EMNLP}
Rush, A.~M.; Chopra, S.; and Weston, J.
\newblock 2015.
\newblock {A Neural Attention Model for Abstractive Sentence Summarization}.
\newblock In {\em Proceedings of EMNLP},  379--389.

\bibitem[\protect\citeauthoryear{Sennrich, Haddow, and
  Birch}{2016}]{sennrich-haddow-birch:2016:P16-11}
Sennrich, R.; Haddow, B.; and Birch, A.
\newblock 2016.
\newblock Improving neural machine translation models with monolingual data.
\newblock In {\em Proceedings of ACL},  86--96.

\bibitem[\protect\citeauthoryear{Shen \bgroup et al\mbox.\egroup
  }{2018}]{shen2018disan}
Shen, T.; Zhou, T.; Long, G.; Jiang, J.; Pan, S.; and Zhang, C.
\newblock 2018.
\newblock Disan: Directional self-attention network for rnn/cnn-free language
  understanding.
\newblock In {\em Proceedings of AAAI}.

\bibitem[\protect\citeauthoryear{Srivastava \bgroup et al\mbox.\egroup
  }{2014}]{Srivastava:2014:DSW:2627435.2670313}
Srivastava, N.; Hinton, G.; Krizhevsky, A.; Sutskever, I.; and Salakhutdinov,
  R.
\newblock 2014.
\newblock Dropout: A simple way to prevent neural networks from overfitting.
\newblock {\em Journal of Machine Learning Research} 15(1):1929--1958.

\bibitem[\protect\citeauthoryear{Sutskever, Vinyals, and
  Le}{2014}]{Sutskever:2014:SSL:2969033.2969173}
Sutskever, I.; Vinyals, O.; and Le, Q.~V.
\newblock 2014.
\newblock {Sequence to Sequence Learning with Neural Networks}.
\newblock In {\em Proceedings of NIPS},  3104--3112.

\bibitem[\protect\citeauthoryear{Takase, Okazaki, and
  Inui}{2016}]{takase-okazaki-inui:2016:P16-1}
Takase, S.; Okazaki, N.; and Inui, K.
\newblock 2016.
\newblock Composing distributed representations of relational patterns.
\newblock In {\em Proceedings of ACL},  2276--2286.

\bibitem[\protect\citeauthoryear{Takase, Suzuki, and
  Nagata}{2017}]{takase-suzuki-nagata:2017:I17-2}
Takase, S.; Suzuki, J.; and Nagata, M.
\newblock 2017.
\newblock Input-to-output gate to improve rnn language models.
\newblock In {\em Proceedings of IJCNLP},  43--48.

\bibitem[\protect\citeauthoryear{Takase, Suzuki, and Nagata}{2018}]{D18-1489}
Takase, S.; Suzuki, J.; and Nagata, M.
\newblock 2018.
\newblock Direct output connection for a high-rank language model.
\newblock In {\em Proceedings of EMNLP},  4599--4609.

\bibitem[\protect\citeauthoryear{Tian, Okazaki, and
  Inui}{2016}]{rantian:additive}
Tian, R.; Okazaki, N.; and Inui, K.
\newblock 2016.
\newblock The mechanism of additive composition.
\newblock {\em CoRR} abs/1511.08407.

\bibitem[\protect\citeauthoryear{Verwimp \bgroup et al\mbox.\egroup
  }{2017}]{verwimp-EtAl:2017:EACLlong}
Verwimp, L.; Pelemans, J.; Van~hamme, H.; and Wambacq, P.
\newblock 2017.
\newblock Character-word lstm language models.
\newblock In {\em Proceedings of EACL},  417--427.

\bibitem[\protect\citeauthoryear{Vinyals \bgroup et al\mbox.\egroup
  }{2015}]{Vinyals_2015_CVPR}
Vinyals, O.; Toshev, A.; Bengio, S.; and Erhan, D.
\newblock 2015.
\newblock Show and tell: A neural image caption generator.
\newblock In {\em Proceedings of CVPR},  3156--3164.

\bibitem[\protect\citeauthoryear{Wan \bgroup et al\mbox.\egroup
  }{2013}]{wan2013regularization}
Wan, L.; Zeiler, M.; Zhang, S.; Cun, Y.~L.; and Fergus, R.
\newblock 2013.
\newblock {Regularization of Neural Networks using DropConnect}.
\newblock In {\em Proceedings of ICML},  1058--1066.

\bibitem[\protect\citeauthoryear{Wen \bgroup et al\mbox.\egroup
  }{2015}]{wen-EtAl:2015:EMNLP}
Wen, T.-H.; Gasic, M.; Mrk\v{s}i\'{c}, N.; Su, P.-H.; Vandyke, D.; and Young,
  S.
\newblock 2015.
\newblock {Semantically Conditioned LSTM-based Natural Language Generation for
  Spoken Dialogue Systems}.
\newblock In {\em Proceedings of EMNLP},  1711--1721.

\bibitem[\protect\citeauthoryear{Wieting \bgroup et al\mbox.\egroup
  }{2016}]{wieting-EtAl:2016:EMNLP2016}
Wieting, J.; Bansal, M.; Gimpel, K.; and Livescu, K.
\newblock 2016.
\newblock Charagram: Embedding words and sentences via character n-grams.
\newblock In {\em Proceedings of EMNLP},  1504--1515.

\bibitem[\protect\citeauthoryear{Yang \bgroup et al\mbox.\egroup
  }{2018}]{DBLP:journals/corr/abs-1711-03953}
Yang, Z.; Dai, Z.; Salakhutdinov, R.; and Cohen, W.~W.
\newblock 2018.
\newblock Breaking the softmax bottleneck: {A} high-rank {RNN} language model.
\newblock In {\em Proceedings of ICLR}.

\bibitem[\protect\citeauthoryear{Zaremba, Sutskever, and
  Vinyals}{2014}]{DBLP:journals/corr/ZarembaSV14}
Zaremba, W.; Sutskever, I.; and Vinyals, O.
\newblock 2014.
\newblock Recurrent neural network regularization.
\newblock In {\em Proceedings of ICLR}.

\bibitem[\protect\citeauthoryear{Zhou \bgroup et al\mbox.\egroup
  }{2017}]{zhou-EtAl:2017:Long}
Zhou, Q.; Yang, N.; Wei, F.; and Zhou, M.
\newblock 2017.
\newblock Selective encoding for abstractive sentence summarization.
\newblock In {\em Proceedings of ACL},  1095--1104.

\bibitem[\protect\citeauthoryear{Zilly \bgroup et al\mbox.\egroup
  }{2017}]{zilly2016recurrent}
Zilly, J.~G.; Srivastava, R.~K.; Koutn{\'\i}k, J.; and Schmidhuber, J.
\newblock 2017.
\newblock {Recurrent Highway Networks}.
\newblock {\em Proceedings of ICML}  4189--4198.

\bibitem[\protect\citeauthoryear{Zolna \bgroup et al\mbox.\egroup
  }{2018}]{fraternal}
Zolna, K.; Arpit, D.; Suhubdy, D.; and Bengio, Y.
\newblock 2018.
\newblock Fraternal dropout.
\newblock In {\em Proceedings of ICLR}.

\bibitem[\protect\citeauthoryear{Zoph and Le}{2017}]{45826}
Zoph, B., and Le, Q.~V.
\newblock 2017.
\newblock {Neural Architecture Search with Reinforcement Learning}.
\newblock In {\em Proceedings of ICLR}.

\end{thebibliography}
\bibliographystyle{aaai}

\end{document}